\pdfoutput=1

\documentclass[11pt]{article}

\usepackage[]{acl}

\usepackage{times}
\usepackage{latexsym}

\usepackage[T1]{fontenc}

\usepackage[utf8]{inputenc}

\usepackage{microtype}
\usepackage{soul}
\usepackage{todonotes}
\usepackage{makecell}

\title{Which anonymization technique is best for which NLP task? - It depends.\\A Systematic Study on Clinical Text Processing}

\author{Iyadh Ben Cheikh Larbi \and  Aljoscha Burchardt \and Roland Roller \\
  Speech and Language Technology Lab \\
  German Research Center for Artificial Intelligence (DFKI) \\
  Alt-Moabit 91c, Berlin, Germany \\
  \texttt{firstname.lastname@dfki.de}  \\}

\begin{document}
\maketitle
\begin{abstract}
%

Clinical text processing has gained more and more attention in recent years. The access to sensitive patient data, on the other hand, 
is still a big challenge, as text cannot be shared without legal hurdles and without removing personal information. 
There are many techniques to modify or remove patient related information, each 
with different strengths. 
This paper investigates the influence of different anonymization techniques on the performance of ML models using  multiple datasets corresponding to five different NLP tasks. Several learnings and recommendations are presented. This work confirms that particularly stronger anonymization techniques lead to a significant drop of performance. 
In addition to that, most of the presented  techniques are not secure against a re-identification attack based on similarity search.


\end{abstract}
\section{Introduction}
While clinical text processing has gained more and more attention in recent years, access to data still remains a major challenge as it typically contains sensitive, patient-related information. A straightforward solution is to apply one of the many existing de-identification and anonymization techniques, and control the access to the data, as for instance in \citet{kittner2021annotation} or \citet{2018}. But each technique has different properties and modification of the source text has effects on machine learning. What happens when you train a model on an anonymized corpus and test it on your own local data (not anoymized)? In which way does this affect the performance of your model?

To put it more generally: If each anonymization technique has different characteristics, how do they affect different natural language processing (NLP) tasks? Is there some rule of thumb we can follow when choosing an anonymization technique, e.g., to share data for a specific task? To explore those questions in detail, this work conducts a systematic analysis regarding the influence of different anonymization techniques and their effects on the performance of (state-of-the-art) machine learning (ML) models. In course of this, we train and test the models using six different datasets corresponding to five different natural language processing tasks. Main contributions of this work are a set of learnings and recommendations regarding text anonymization for NLP tasks, as well as a small, fictitious re-identification experiment to explore the (in-)effectiveness of the different techniques. The software implementations of the different anonymization techniques will be made publicly available.






\section{Related Work}

In accordance with the HIPAA Safe Harbor \cite{HIPAA} method, we define de-identification as the removal of protected health information (PHI) 
that directly relate to an individual such as name, address, birth date, etc. However, de-identification does not guarantee anonymity for data subjects. 
Anonymization on the other hand is defined as any irreversible procedure, in which no information can be linked to any individual \cite{nolinkage}, making the data subjects anonymous and no longer identifiable. 

A range of different text anonymization approaches exist in the literature, which modify the text structure within a dataset, delete, replace, or introduce synthetic information, to make it harder to identify or infer factual information on the patient. The following approaches have been explored for this work: 

\begin{itemize}
    \item \textbf{\textit{Suppression}} \cite{mamede2016automated} is a technique that either completely removes certain words or sentences or masking them with a neutral label denoting their suppression.
    \item \textbf{\textit{Perturbation}} \cite{zuo2021data} modifies data through permutation or data swapping, in case of text, similarly to data augmentation, by flipping characters, or changing the order of words.
    \item \textbf{\textit{Substitution}} \cite{mamede2016automated} replaces certain information with more general terms.
    \item Finally, \textbf{\textit{Aggregation}} (k-anonymity) \cite{samarati1998protecting} groups individual data subjects together, e.g. by their attribute values, to make it more difficult to identify a single individual.
\end{itemize}

Only limited work has been done to describe the (systematic) influence on text anonymization on the performance of ML models, most work targets de-identification only. \citet{meystre2014text} for instance examine evaluates information loss after de-identification by examining the text-level changes rather than the performance of the machine learning models. The work concludes that only 1.2-3\% of the clinical concepts are changed by de-identification and that the overall impact on the clinical information is minimal but not negligible.

The work of \citet{obeid2019impact} analyzes the impact of de-identification on a binary classification task, and concludes, that there is no significant difference in performance between training on the original texts and training on the de-identified texts. Similarly, \citet{berg2020impact} examines the effect of different PHI concealment strategies on named entity recognition (NER) tasks and show that using moderate to high precision de-identification models with the right concealment strategy leads to similar performance. Furthermore, 
\citet{vakili2022downstream} explored the effects of two approaches to de-identification, namely pseudonymizing PHI in a text and removing PHI-including sentences. The work concludes, that there is no negative impact on the performance of the models on downstream NLP tasks such as NER, text classification, etc.   

Also the work of \citet{lange2020closing} explore the performance of concept extraction using de-identified data and conclude the performance drop is only marginal. Finally, although not clinical text, in \citet{impactsentiment} show that anonymization can cause significant negative changes in the sentiment analysis performance on Twitter data.

This work however goes beyond existing related work, as we carry out a structured analysis regarding anonymization of clinical text, testing seven different techniques, on six datasets, including five different NLP tasks.

\section{Data and Methods} \label{Data and Methods}

\begin{table*}[h!]
\centering
\scriptsize
\begin{tabular}{| c | c | c | c | c | c | c | c | c | c | c | c |}
\hline
\textbf{Corpus} &         \textbf{DeI} &      \textbf{MNr} &     \textbf{ShS} &      \textbf{RaS 20\%} &      \textbf{RaS 100\%} &      \textbf{SyR 20\%} &      \textbf{SyR 100\%}  &        \textbf{CnR} &        \textbf{Ag2} &        \textbf{Ag3} &        \textbf{Ag4} \\
\hline
\textbf{Smoking} &  +1.43 &  +0.27 &  +1.05 &  -5.09*  &  -5.46*  &  -4.74 &   -8.31* &  +0.22 &  -6.34* &    -6.80* &   -7.25*  \\
\textbf{Obesity} &   +0.80  &  -0.61  &  -2.55*  &  -1.94*  &   -5.09*  &  -2.99*  &   -8.96*   &  -1.31*  &  -12.48*  &  -22.59*  &  -36.97*  \\
\textbf{MedNLI}  &  +1.55*  &  +0.14  &           - &  -1.13  &   -1.93*  &  -2.52*  &   -8.42*   &  -0.73  &   -7.98*  &  -13.34*  &  -14.81*  \\
\textbf{ClinSTS} &  -1.21  &  -0.12  &           - &  -1.36  &   -0.95  &  -1.92  &  -21.96*   &  -1.84*  &    -3.30*  &   -7.26*  &  -24.31*  \\
\textbf{2010}    &  -0.32  &   -0.50*  &           - &  -4.34*  &  -16.94*  &  -5.96*  &  -15.77*   &  -2.48*  &            - &            - &            - \\
\textbf{2018}    &  -0.83  &   -5.10*  &           - &  -3.04*  &  -25.12*  &  -2.73*  &   -9.72*   &  -1.19*  &            - &            - &            - \\
\hline
\textbf{mean}    &     +0.368 &     -0.855 &     -0.355 &     -2.692 &     -3.353 &      -8.907 &     -12.232  &     -1.092 &      -7.342 &     -12.315 &     -20.655 \\
\hline
\end{tabular}
\caption{Anonymization Effects: Average performance drop/gain across all runs in percent in comparison to the best performing system on the corresponding task, according to Table \ref{table:BenchmarkResults}. Significant (p<0.05) results are marked with *}
\vspace{-4mm}
\label{table:AnonymizationEffects}  
\end{table*}

The experiments in this work are based on the following   datasets and tasks: 

\begin{itemize}
    \item \textbf{2010 i2b2/VA} \cite{2010} (named entity recognition, NER)
    \item \textbf{2018 n2c2} \cite{2018} (named entity recognition, NER)
    \item  \textbf{2006 Smoking Challenge} \cite{2006} (multi-class classification, MCC)
    \item \textbf{2008 Obesity Challenge} \cite{2008} (multi-label classification, MLC)
    \item \textbf{MedNLI} \cite{mednli} (natural language inference, NLI)
    \item \textbf{ClinSTS} \cite{clinsts} (semantic textual similarity, STS)
\end{itemize}

While the first four datasets include annotated discharge summaries, the last two datasets include pairs of sentences extracted from MIMIC-III \cite{mimic}. Due to limited space, we refer the reader to the source papers and to the appendix.


Using those datasets, different text anonymization techniques are applied to the training split. The following techniques are used, based on Suppression, Perturbation, Substitution and Aggregation, as described above: 

\begin{enumerate}
    \item \textbf{\textit{De-identification (DeI)}} Using the tool Philter \cite{philter}, all PHI data in the text is replaced by "XXXX". 
    \item \textbf{\textit{Mask Numbers (MNr)}} All occurrences of numbers in a given text, both in numerical or alphabetical form, are replaced using ``XX''.
    \item \textbf{\textit{Shuffle Sentences (ShS)}} Sentences in a given text are shuffled.
    \item \textbf{\textit{Random Swap (RaS)}} A certain percentage of words are randomly chosen and swapped all over the document.
    \item \textbf{\textit{Synonym Replacement (SyR)}} A certain percentage of the non-stop words in the document are replaced with WordNet synonyms.
    \item \textbf{\textit{Clinical Concept Synonym Replacement (CnR)}} All signs/symptoms, diseases/disorders, and medications are replaced by a random UMLS synonym, using cTAKES \cite{ctakes} for entity linking. 
    \item \textbf{\textit{Text Aggregation (AgX)}} is done by merging a certain amount of shuffled documents (X) into one. 
\end{enumerate}

Finally, in order to examine the effect of anonymization on the performance of state-of-the-art machine learning models, we rely on existing BERT solutions which have achieved promising results on the different datasets in the recent past. Particularly we rely on \textbf{BERT base (uncased)} \cite{bert}, \textbf{Bio+Clinical BERT} \cite{bioclin}, as well as \textbf{BERT long document classification} \cite{bert_doc_cls}. 

\section{Experiments}

\subsection{Setup}

For our experiments, we rely - if possible - on the original setup and configuration as described in the original publications. Given the clinical corpus, the data is split into training and test data. 
Next the anonymization is applied to the training data. Together with each anonymization, a model is trained and then evaluated on the original (not anonymized) text of the test split. For each technique, the model is trained and evaluated five different times. If the anonymization technique is not deterministic and produces a different anonymized dataset each time, we repeat the text anonymization five times, which results in 25 runs. The results of each approach are averaged and compared to the performance of the base model (without anonymization).

All experiments are conducted with \textbf{BERT base} and \textbf{Bio+Clinical BERT}. The experiments corresponding to the classification tasks (Smoking and Obesity), are additionally conducted with \textbf{BERT long document classification}, as documents in those tasks are quite long. In case of random swap, random replacement the presented anonymization is applied to 20 and 100\% of the data.

\subsection{Results}

First, each model has been trained and tested on the original data - without applying the anonymization beforehand. Results are presented in Table \ref{table:BenchmarkResults}. Note, our base results slightly differ from the results in the ref. papers, even though we use the same setup. 

\begin{table}[!h]
\centering
\tiny
\begin{tabular}{| c | c | c | c | c | c | c |}
\hline
\textbf{Model} & \textbf{Smoking} & \textbf{Obesity} & \textbf{MedNLI} & \textbf{ClinSTS} &   \textbf{2010} &   \textbf{2018} \\
\hline
\textbf{BERT} &   77.89 &   67.58 &   76.9 &   83.88 &  82.62 &  87.84  \\
\textbf{BioC} &   75.48 &   70.73 &  \textbf{80.49} &   \textbf{84.83} &  \textbf{84.54} &  \textbf{89.03}  \\
\textbf{LDoc} &   \textbf{87.69} &   \textbf{82.51} &      - &       - &      - &      -  \\
\hline
\textbf{Eval} & F1 & F1 & Acc. & Pearson & F1 & F1\\
\hline
\end{tabular}
\caption{Base results on all datasets in terms of average scores across all runs, using BERT base, Bio+Clinical BERT (BioC) and BERT long document classification.}
\vspace{-2mm}
\label{table:BenchmarkResults}  
\end{table}

Next we apply the different text anonymization techniques to the training data, train the models and test them on the original data. The results of the different techniques, in comparison to the best performing base system on that task, are presented in Table \ref{table:AnonymizationEffects}.


\begin{table*}[h!]
\centering
\scriptsize
\begin{tabular}{| c | c | c | c | c | c | c | c | c | c | c | c |}

\hline

{} &     \textbf{DeI} &  \textbf{MNr} & \textbf{ShS} &  \textbf{RaS 20\%} &  \textbf{RaS 100\%} & \textbf{ SyR 20\%} & \textbf{SyR 100\%} &   \textbf{CnR} &    \textbf{Ag2} &    \textbf{Ag3} &    \textbf{Ag4} \\
\hline
\textbf{found}               &     1.0 &     1.0 &     1.0 &     1.0 &     1.0 &     1.0 &     1.0  & 1.0 &  0.9063 &  0.6351 &  0.2789 \\
\textbf{a/o sim} &  0.9529 &  0.8949 &     1.0 &  0.9986 &  0.9986 &  0.5486 &  0.2442  & 0.7512 &  0.5758 &  0.4261 &   0.3470 \\
\textbf{avg-sim} &  0.1502 &  0.1458 &  0.1524 &  0.1518 &  0.1518 &  0.1084 &  0.0589  & 0.1388 &  0.1646 &  0.1603 &  0.1531 \\
\hline

\end{tabular}
\caption{Re-identification of patients using different text anonymization techniques. \textit{found} refers to the percentage of cases in which the highest ranked (most similar) document on the original dataset was the correct one; \textit{a/o sim} describes the distance between the anonymized document and its original version; \textit{avg-sim} describes the average similarity between a given anonymized document all 3500 original documents}
\vspace{-4mm}
\label{table:Reidentification}  
\end{table*}

\subsection{Analysis}

The conducted suppression methods \textit{de-identification} (DeI) and \textit{mask number} (MNr), mask some information with a neutral label (`XXXX'). In most cases the general effect is rather minimal. Particularly in the case of \textit{DeI}, the table shows a slight improvement of performance. This is surprising, and might be connected to the random model initialization. 
Another reason could be that, from a model perspective, less relevant information has been discarded. However, only in the case of \textbf{MedNLI} the results are significantly better. \textit{MNr} on the \textbf{2018} task causes a moderate performance loss due to entities related to numerical values, such as dosage or strength. Overall, the results are inline with the findings presented by \citet{berg2020impact} and \citet{lange2020closing}.


In our experiment, perturbation changes the sentence order (\textit{sentence shuffle}; \textit{ShS}) and the order to the words within the document (\textit{random swap}; \textit{RaS}). Different from suppression, the technique shows a stronger performance loss, particularly in case of \textit{RaS}. The more words swapped across the document, the stronger in most cases the drop of performance (RaS 20\% versus 100\%). The technique has a particularly strong influence on NER tasks, in which the word order plays an important role. 
Instead, using \textit{sentence shuffle} a negative significant effect can be observed on the obesity task.   


Similarly it behaves with the substitution techniques (WordNet) \textit{synonym replacement} and (UMLS) \textit{clinical concept synonym replacement}. Generally both techniques lead to a drop of performance, which is stronger the more words affected by the technique (applying to 20\% of the data versus 100\%). The drop is notably stronger in case of \textit{synonym replacement}, as more words are affected and possibly also wrong synonyms might have been inserted - depending on the context. In case of \textit{clinical concept synonym replacement}, the performance loss is notably smaller, as possibly less words are affected. Also, according to the frequency of UMLS mentions, in various cases the preferred concept mention might have been chosen. 

Finally, text aggregation, which merges documents according to different characteristics, has the strongest effect on the model performance. For all tasks we can observe, the more files are aggregated the stronger the drop in performance. We stopped with a maximum of 4 documents (Ag4), as the document length of the merged case reports was too long otherwise. In case of text classification, NLI and STS documents with the same labels have been merged together, thus the effect might not be too strong. However, in case of multi-label classification (Obesity) the new aggregated documents are now not only larger, but also contain more labels. 


\subsection{Re-Identification Experiment}
\label{sec:bibtex}

The previous experiment presented the influence of each anonymization technique on the different NLP tasks. To validate the efficiency and robustness of each technique, we conduct in the following a small and simple fictitious re-identification experiment. The question we investigate is, how difficult it would be, to link anonymized text to a particular patient, assuming that an attacker has the anonymized text and access to the original patient database. We conduct this small experiment using 3500 texts from MIMIC-III. The setup is as follows: First we run the different anonymization techniques on the data, and then we start a similarity search by calculating the Jaccard Distance on word level, between each anonymized document and all (original) 3500 MIMIC texts. 

Our setup describes a worst-case scenario, and we hope that it is unlikely to happen. However the scenario describes how much the anonymized document differs from its original version, and how easily the original could be found using a simple word based similarity search. As depicted in Table \ref{table:Reidentification}, the average similarity (avg-sim) from an anonymized document to the documents in the MIMIC dataset is mostly about \textit{0.15}. Instead the similarity to the correct document (a/o-sim) is always above this average score. However, while in case of suppression and perturbation techniques the a/o-sim score is about \textit{0.9--1}, the similarity strongly decreases with substitution and aggregation, most notably with SyR 100\% and Ag4. Conversely, only in case of aggregation the highest ranked documents are not necessarily the corresponding original documents, thus providing some (minor) security against a possible re-identification in our scenario. Based on the outcomes we define an anonymization as `stronger', the lower the values \textit{a/o sim} and \textit{found} are.


\section{Recommendations and Learnings}


Based on the outcomes of the previous two experiments, we draw the following conclusions regarding clinical text anonymization:



There is no one-size-fits-all anonymization technique that can always be recommended. The optimal technique needs to be selected depending on the (security) requirements, the sensitivity of the data as well as underlying NLP task. Overall, the results indicate a correlation between performance loss and strength of anonymization technique, but each technique of course comes at a cost. While some can be quickly conducted, such as sentence shuffle or aggregation, others require additional tools and resources such as \textit{DeI} or \textit{SyR}, which can prohibit their use in some scenarios. 

Text aggregation is the strongest of the presented techniques. It offers relatively good security against re-identification, but leads to the strongest performance loss. This technique not only aggregates the texts and their contexts together but also results in less training data, which is one of the reasons for the performance loss, as seen in the appendix. Although text aggregation is generally the technique of choice for providing the maximum security, in case of multi label classification, it suffers a strong performance drop. In this case, we recommend relying on substitution techniques, such as synonym replacement. To provide further abstraction, substituted data could be shuffled and be possibly enriched with additional sentences. 

Another disadvantage of text aggregation is the fact that long text documents get even longer. Depending on the NLP task, documents need to be processed at once (e.g. document classification) and standard BERT models can deal only with up to 512 input tokens. Relevant information might get lost, and models result in lower performance.

Tasks which can be conducted on sentence level can be easier detached from a patient, and thus weaker anonymization techniques can be applied. Finally, the applied perturbation techniques appear not to be useful for the given tasks due to the weak anonymization.







\section{Conclusion}

This work presented a structured analysis regarding text anonymization and its influence on the machine learning model performance. Our experiment tested seven different anonymization techniques on multiple datasets, including five different clinical NLP tasks. In addition, we conducted a simple fictitious re-identification experiment to examine the robustness of each technique. Together with the results, we present some recommendations and learnings. For short, we did not find a one-size-fits-all anonymization technique that would perform best in all tasks. The particular decision depends on several factors. In addition, we provide the software to conduct the text anonymization studies. In future work, more de-identification and anonymization techniques could be added to hopefully arrive at a comprehensive overall picture.

\bibliography{references} 

\appendix

\begin{table*}[h!]
\centering
\small
\begin{tabular}{|c||m{1.7cm}|m{1.7cm}|m{1.7cm}|m{1.7cm}|m{1.7cm}|m{1.7cm}|}
\hline 
 \textbf{Dataset Name}  & \textbf{Smoking} & \textbf{Obesity} & \textbf{MedNLI} & \textbf{ClinSTS} & \textbf{2010} & \textbf{2018} \\
\hline 
  \textbf{NLP Task} & multi class classification (MCC) & multi label classification (MLC) & natural language inference (NLI) & semantic textual similarity (STS) & named entity recognition (NER) & named entity recognition (NER)  \\
\hline 
 \textbf{Train size} & 389 & 730 & 11232 & 1642  & 170 & 303 \\
\hline 
 \textbf{Dev size} & 104* & 507* & 1395 & 412*  & 256* & 202* \\
\hline 
 \textbf{Test size} & 104 & 507 & 1422 & 412  & 256 & 202 \\
\hline 
 \textbf{Type of data} & discharge summaries  & discharge summaries & sentence pairs & sentence pairs  & discharge summaries** & discharge summaries** \\
\hline 
 \textbf{Avg \# token} & 1100.61 & 1935.72  & 37.22 & 57.65  & 15.41 & 63.657\\
\hline 
\end{tabular}

\caption{Overview about dataset used for the anonymization experiments. (*) For all datasets except MedNLI no development (dev) set was provided. Therefore, the test set was also used as a dev set instead. (**) The discharge summaries of the NER tasks have been divided into individual sentences which have been used to train and test the models. The anonymization techniques were directly applied on those sentences.}
\label{table:Datasets overview}     
\end{table*}

\section{Appendix}
\label{sec:appendix}

\subsection{Overview about data}

\noindent

In total, six text-based clinical datasets have been chosen in this work to examine the effects of anonymization on their respective tasks.
Table \ref{table:Datasets overview} provides an overview on these datasets as well as some details about their sizes and types.


\subsection{Text anonymization techniques - some more details}

Here some additional details we left out of the main article, due to limited space:

\paragraph{De-identification} In all datasets we used for our experiments were already de-identified (PHI information was replaced with pseudonyms). However, we tested de-identification in our experiments, as we wanted to explore the influence of masking PHI information. 

\paragraph{Percentage} The two techniques of random swap and synonym replacement apply the technique to a `certain percentage of (non-stop) words'. We have tested those techniques with different number of modification steps (in percentage), however, in the main article we just report 20\% and 100\%.

\paragraph{Clinical Concept Synonym Replacement} The technique replaces each detected signs/symptoms, diseases/disorders, and medications with a synonym provided by UMLS. This synonym is selected randomly from all English mentions corresponding to the given concept unique identifier (CUI). This means that also the previously given entity mention could be again inserted by this random selection. In this way, more common mentions are favored over seldom entity mentions in UMLS.

\paragraph{Aggregation} Text aggregation merges two or more files together into one file. In this way the size and number of tokens is significantly increased which makes it harder for standard BERT models (512 tokens) on discharge summaries. For this reason we also used BERT long document classification. Moreover, text aggregation decreases the size of the training data by the factor of files which are merged together. Merging for instance four files into one, decreases the data to 25\% of its original size. Repeating the aggregation step several times with different files, would create a larger dataset, but makes it much easier to re-identify single patient documents. Thus, the aggregation would have no effect anymore. See more in Section \ref{AugTextAgg}.

Text aggregation merges documents according to their target label. In case of the \textit{smoking} task for instance, only documents reporting the exact same smoking status could be merged together (randomly). However, this did not work in case of multi label classification. Thus, in case of \textit{obesity} random documents were aggregated.

\subsection{Implemented Models}
To perform the experiments, we implemented six different models where the pre-trained BERT models (BERT Base uncased and Bio+Clinical BERT) can be adapted and fine-tuned to perform the tasks behind the corresponding datasets. The configuration and hyper-parameters  of the models are presented in Table \ref{tableModelsOverview}.

\begin{table*}[h!] 
        \centering

\small
\begin{tabular}{|c||m{1.5cm}|m{1.5cm}|m{1.5cm}|m{1.5cm}|m{1.5cm}|m{1.5cm}|}
\hline 
 \textbf{Models} & \textbf{Smoking} & \textbf{Obesity} & \textbf{MedNLI} & \textbf{ClinSTS}  & \textbf{2010} & \textbf{2018}\\
\hline 
\hline 
 \textbf{Token sequence length} & 512 & 512  & 150 & 150  & 150 & 200\\
\hline 
 \textbf{Epochs} & 40* & 40* & 3 & 3 & 3 & 3 \\
\hline 

\textbf{Linear layer's input} & [CLS] token's output embedding  & [CLS] token's output embedding  & [CLS] token's output embedding & [CLS] token's output embedding & Each token's output embedding & 
Each token's output embedding \\

\hline 
 \textbf{Linear layer's input size} & 768 & 768 & 768 & 768 & 768 & 768 \\
\hline 
 \textbf{Linear layer's output size} & 5 & 16 & 3 & 1 & 9 & 22  \\
\hline 
 \textbf{Loss function} & Cross entropy loss & Binary cross entropy loss & Cross entropy loss & Mean squared error  & Cross entropy loss & Cross entropy loss \\
\hline 
 \textbf{Activation function} & Softmax & Sigmoid  & Softmax & Identity function & Softmax & Softmax \\
\hline 
\textbf{Evaluation function} & Micro-F1 & Micro-F1 & Accuracy & Pearson  & Micro-F1 & Micro-F1\\
\hline 
 \textbf{Optimizer} & AdamW & BertAdam & BertAdam & BertAdam & BertAdam & BertAdam \\
 \hline
\end{tabular}

 \caption{Models overview.}
\label{tableModelsOverview}    
\end{table*}

In addition to the implemented models, the BERT long document classification model has been downloaded from github \footnote{\url{https://github.com/AndriyMulyar/bert\_document\_classification}}
and used as is on the text classification datasets and their anonymized versions. Some preprocessing might be necessary to input the anonymized data to the model but the architecture is intact. 
Furthermore, during the preprocessing of the NER texts, the words have been annotated according to the IOB2 format.

\subsection{Augmented Text Aggregation}
\label{AugTextAgg}

Among the anonymization techniques tested during this work is one called \textbf{\textit{Augmented Text Aggregation (AAgX)}}. It consists in repeating the simple Text Aggregation technique presented in Section \ref{Data and Methods}, $n$ independent times. The resulting datasets are then simply merged together. We introduced this technique to generate more training data as the Text Aggregation technique reduces the training size to a factor of $1/n$.
This should not be used in practice as it makes it easier, with more resources, to single out patients that are present in multiple texts. 

Table \ref{table:AugmentedTextAggregation} highlights the performance drops or gains after applying this anonymization techniques to the datasets, in comparison to the reported results in Table \ref{table:AnonymizationEffects}. In all cases, the results of the Augmented Text Aggregation are better than those of the simple Text Aggregation, most notably for ClinSTS. 

\begin{table}[h!]
\centering
\small
\begin{tabular}{| c | c | c | c |}
\hline
{} &     \textbf{AAg2} &      \textbf{AAg3} &      \textbf{AAg4} \\
\hline
\textbf{Smoking} & +2.89 &   -4.88 &  -6.25  \\
\textbf{Obesity} &  -5.84  &  -14.48  &  -31.68  \\
\textbf{MedNLI}  &   -6.60  &  -11.09  &  -12.13  \\
\textbf{ClinSTS} &  -1.45  &   -3.04  &   -3.14  \\
\textbf{2010}    &           - &            - &            - \\
\textbf{2018}    &           - &            - &            - \\
\hline
\textbf{mean}    &      -2.55 &      -8.19 &     -13.12 \\
\hline

\end{tabular}
\caption{Augmented Text Aggregation}
\label{table:AugmentedTextAggregation}  
\end{table}

\end{document}